\documentclass[letterpaper, 10 pt, conference]{ieeeconf}  % Comment this line out if you need a4paper
\IEEEoverridecommandlockouts

\usepackage{amsmath,amssymb,amsfonts}
\usepackage{graphicx}
\usepackage{color,soul} %\sout
\definecolor{red}{rgb}{1.00,0.00,0.00}
\definecolor{blue}{rgb}{0.00,0.00,1.00}
\definecolor{green}{rgb}{0.30, 0.50,0.00}

\definecolor{darkgreen}{rgb}{0.0, 0.4, 0.0}
\usepackage{threeparttable} % For adding notes below the table
\usepackage{pgf}
\usepackage{cite}
\usepackage{wrapfig}
\usepackage{algorithmic}
\usepackage{textcomp}
\usepackage{xcolor}
\usepackage{hyperref}
\usepackage{booktabs}
\usepackage{xcolor} % Required for color support
\usepackage{diagbox}
\usepackage{tabularx}
\usepackage{multirow}
\usepackage{dcolumn}
\usepackage{makecell}
\definecolor{lightgreen}{RGB}{144,238,144}
\definecolor{lightred}{RGB}{255,182,193}
\def\BibTeX{{\rm B\kern-.05em{\sc i\kern-.025em b}\kern-.08em
    T\kern-.1667em\lower.7ex\hbox{E}\kern-.125emX}}

\begin{document}

\title{LM-MCVT: A Lightweight Multi-modal Multi-view Convolutional-Vision Transformer Approach for 3D Object Recognition}

\author{Songsong Xiong$^{1}$, Hamidreza Kasaei$^{1}$% <-this % stops a space
% \thanks{*This work was not supported by any organization}% <-this % stops a space
\thanks{$^{1}$Department of Artificial Intelligence,
        University of Groningen, Groningen, The Netherlands\newline 
        {\tt\small \{s.xiong, hamidreza.kasaei\}@rug.nl}}}

\maketitle
\thispagestyle{empty}
\pagestyle{empty}

%%%%%%%%%%%%%%%%%%%%%%%%%%%%%%%%%%%%%%%%%%%%%%%%%%%%%%%%%%%%%%%%%%%%%%%%%%%%%%%%
\begin{abstract}
In human-centered environments such as restaurants, homes, and warehouses, robots often face challenges in accurately recognizing 3D objects. These challenges stem from the complexity and variability of these environments, including diverse object shapes. In this paper, we propose a novel Lightweight Multi-modal Multi-view Convolutional-Vision Transformer network (LM-MCVT) to enhance 3D object recognition in robotic applications. Our approach leverages the Globally Entropy-based Embeddings Fusion (GEEF) method to integrate multi-views efficiently. The LM-MCVT architecture incorporates pre- and mid-level convolutional encoders and local and global transformers to enhance feature extraction and recognition accuracy. We evaluate our method on the synthetic ModelNet40 dataset and achieve a recognition accuracy of $95.6\%$ using a four-view setup, surpassing existing state-of-the-art methods. To further validate its effectiveness, we conduct 5-fold cross-validation on the real-world OmniObject3D dataset using the same configuration. Results consistently show superior performance, demonstrating the method’s robustness in 3D object recognition across synthetic and real-world 3D data.

\end{abstract}
\vspace{0.2em}

%%%%%%%%%%%%%%%%%%%%%%%%%%%%%%%%%%%%%%%%%%%%%%%%%%%%%%%%%%%%%%%%%%%%%%%%%%%%%%%%

\section{Introduction}
As societal growth accelerates, the increasing labor shortages are driving the integration of robots into human-centered environments such as homes, warehouses, and factories. These environments demand robots capable of performing tasks with efficient and accurate object recognition~\cite{liu2023lightweight}. Effective object perception enables robots to understand, and safely interact with their surroundings. In recent years, advancements in three-dimensional (3D) object recognition technologies have significantly enhanced robots' capabilities in perception, particularly in assistive and service-oriented robotic applications.

\begin{figure}[!t]
\vspace{1.5mm}
\centerline{\includegraphics[width=\linewidth]{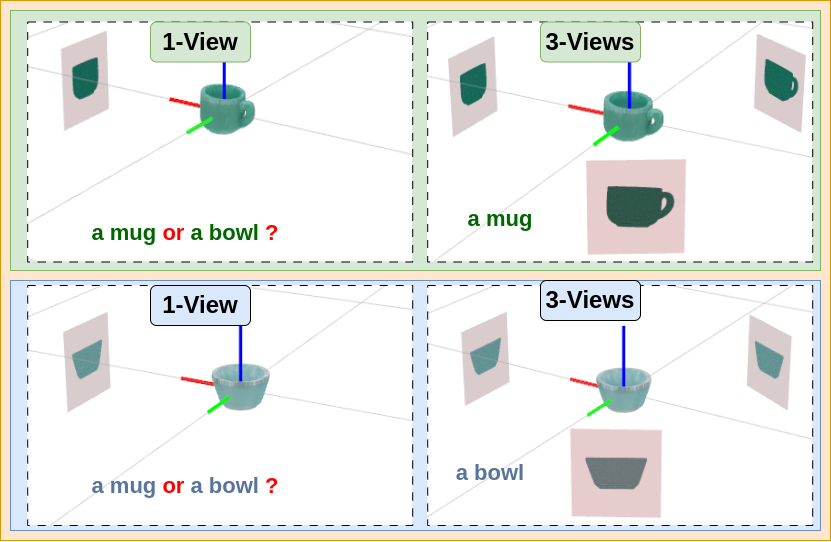}}
\caption{An illustrative example of enhancing 3D object recognition by considering multiple viewpoints: In a single-view setup, different objects may look very similar and cannot be distinguished. In this example, the robot might be confused about whether the object is a \textit{bowl} or \textit{mug}. However, by considering more views of the object, the robot can observe specific features of the object and improve its recognition accuracy.}
\label{robot1}
\vspace{-5mm}
\end{figure}
In general, 3D object recognition approaches can be categorized into three main categories: voxel-based, point-based, and view-based methods. Voxel-based methods~\cite{maturana2015voxnet, putra2023fuzzy,he2024similarity} and point-based methods~\cite{qi2017pointnet, wang2024pointramba, pang2022masked, zhang2024point} use 3D meshes and point clouds, respectively. While these approaches comprehensively capture the spatial information of 3D objects, their high computational costs limit practicality~\cite{zhou2019multi}. In contrast, view-based methods enhance efficiency by converting 3D objects into 2D images from various viewpoints, simplifying the process and eliminating complex 3D feature processing~\cite{qi2021review}. For example, as shown in Fig.~\ref{robot1}, a mug with rich features can be accurately recognized from multiple viewpoints. Benefiting from the advancement of deep learning, various methods encoded the rendered views to achieve a robust representation of a given object. For instance, ~\cite{kasaei2024simultaneous,Alzahrani_2024_CVPR, wang2022ovpt} utilize 2D CNNs to extract features from multiple views of an object and aggregate these features for 3D object classification purposes.  However, CNNs primarily focus on capturing local details while often overlooking global contextual relationships. Inspired by the ability of vision transformers to capture global spatial information, ~\cite{Chen2021MVT} introduced the multi-view vision transformer (MVT) for enhanced 3D object recognition. However, MVT relies on a simple average-pooling operation for view aggregation, which overlooks variations and unique contributions among views. Furthermore, as the number of ViT encoders increases, the model's ability to retain local features weakens, limiting its overall performance. Previous research on multi-view object recognition was mainly focused on dense multi-view techniques to achieve better accuracy. Increasing the number of views can, to some extent, improve recognition accuracy, but acquiring dense multi-view data in real-world applications is often constrained by computational and memory requirements. Additionally, single-view methods incorporating depth information have been introduced to enhance object representation for 3D object recognition~\cite{tziafas2023early}. Compared to dense multi-view approaches, these single-view methods offer better efficiency but still face limitations in stability and robustness~\cite{tziafas2023early}.

\begin{figure*}[!t]
\vspace{3.5mm}
\includegraphics[width=\linewidth]{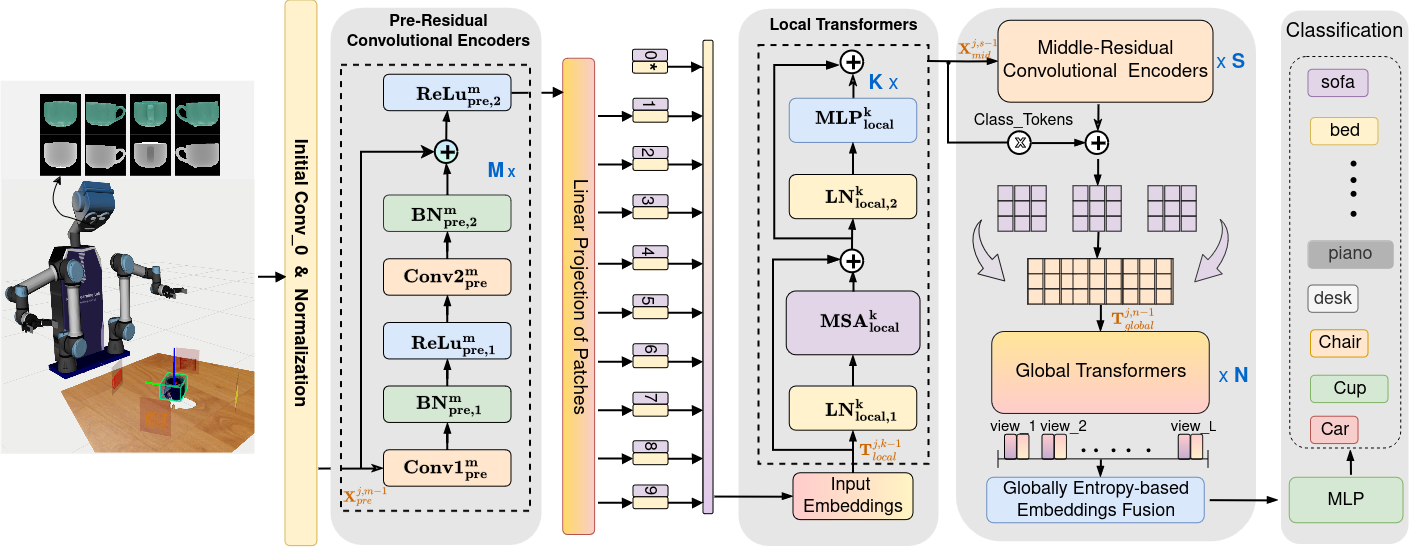}
\caption{LM-MCVT Framework for 3D Object Recognition in Robotic Perception. The 3D object, i.e., {\textit{\textbf{a cup}}}, is projected onto RGB and Depth images from multiple viewpoints. After initial convolution and normalization, the multi-view representations are encoded using \textit{\textbf{\textcolor{blue}{M}}} x pre-residual convolutional encoders and projected into patches. These undergo global feature learning via~\textit{\textbf{\textcolor{blue}{K}}} x local transformers and are refined through~\textit{\textbf{\textcolor{blue}{S}}} x middle-residual encoders. Class token embeddings are combined with middle-residual output in~\textit{\textbf{\textcolor{blue}{N}}} x global transformers. Finally, global entropy-based fusion integrates the multi-view representations for 3D object recognition.}
\vspace{-2mm}
\label{fgvc_overview}
\end{figure*}

To achieve robust and efficient 3D object recognition, we propose the Lightweight Multi-modal Multi-view Convolutional-Vision Transformer network (LM-MCVT), illustrated in Fig.~\ref{fgvc_overview}. We conduct extensive experiments, showing that our method surpasses state-of-the-art approaches in 3D object recognition performance. Furthermore, we deploy our model in real-world robotic scenarios for 3D object recognition and manipulation, demonstrating its robustness and reliability. In summary, our key contributions are twofold:

\begin{itemize}
\item We propose a Lightweight Multi-modal Multi-view Convolutional-Vision Transformer (LM-MCVT) network for 3D object recognition. Our approach integrates convolutional encoders and vision transformers to jointly capture local and global features, achieving superior performance with significantly reduced computational cost compared to existing methods.

\item We develop the Globally Entropy-based Embeddings Fusion (GEEF) method, a novel feature aggregation strategy that leverages entropy-based weighting to adaptively fuse embeddings from multiple viewpoints. This approach efficiently integrates multi-modal multi-view representations, enhancing the performance of 3D object recognition.
\end{itemize}

\section{Related work}

In this section, we review the key efforts in 3D object recognition, focusing on voxel-based, point-based, and view-based methods.

\noindent \textbf{Voxel-based methods}: Such approaches represent an object's point cloud data using discretized 3D voxel structures. For example, Maturana et al.~\cite{maturana2015voxnet} introduced the VoxNet framework for 3D object classification, utilizing an integrated volume-occupying grid representation with 3D CNNs. However, the performance of VoxNet is constrained by its resolution due to sparse data~\cite{wang2022multi}. He et al.~\cite{he2024similarity} combined spatial-filling curves with the octree structure to encode volumetric data. Although this approach reduces memory usage compared to voxel grids, it still requires significant memory for large-scale or high-resolution 3D data.

\noindent \textbf{Point-based methods}: These approaches use point cloud data directly. For instance, PointNet~\cite{qi2017pointnet}, a pioneering method, recognizes 3D objects by directly processing point cloud data. 
To enhance the performance of PointNet, MAP~\cite{pang2022masked}, a transformer-based approach, was developed to extract high-level latent features from unmasked point patches. Point cloud mamba was introduced to model point cloud data at a global level, further improving performance~\cite{zhang2024point}.
Compared to voxel-based methods, point-based approaches efficiently handle unstructured and irregularly sampled data, avoiding the need for uniform grids or meshes. However, accurate recognition in point-based methods still requires substantial computational resources, limiting their practical applicability~\cite{zhou2019multi}.

\noindent \textbf{View-based methods}: These methods represent a 3D object by projecting it into multiple 2D images captured from various viewpoints. Our method belongs to this category. Hou et al.~\cite{hou2024learning} used CNNs to encode multiple views, then applied reinforcement learning to select the next-best-view for 3D object recognition.
Alzahrani et al.~\cite{Alzahrani_2024_CVPR} employed the most discriminative views of an object and achieved competitive accuracy in 3D object classification tasks. While these approaches have advanced the field, their reliance on heavy backbone networks often results in high memory usage and increased computational time, posing challenges for mobile devices. Moreover, such methods frequently prioritize extracting local features over capturing global spatial information~\cite{raghu2021vision}. Chen et al.~\cite{Chen2021MVT} proposed the Multi-view Vision Transformer (MVT), a 3D object classification network that uses the middle-weight DeiT backbone. MVT incorporates multi-head attention to capture global object information and enables effective communication across different views through a unified global structure. However, as the number of layers in the MVT backbone increases, the focus on capturing local details from input views diminishes~\cite{raghu2021vision}. To further enhance 3D object recognition, Wang et al. proposed the optimal viewpoint pooling transformer, which obtains the best viewpoint settings from dense views to optimize performance~\cite{wang2022ovpt}. In contrast to our approach, this method faces substantial challenges when optimal viewpoints are unavailable. Obtaining all views or dense views for a mobile robot in real-world scenarios, particularly in dynamic environments, presents practical difficulties.

Compared to the previous RGB-only methods, RGBD approaches demonstrate superior performance, especially when fewer views are available. For example, Kumra et al.~\cite{kumra2020antipodal} introduced a RGBD method to effectively guide robotic manipulation. Tziafas et al.~\cite{tziafas2023early} transferred pre-trained Vision Transformers to the RGBD domain for 3D object recognition by focusing on late-fusion cross-modal interactions at the downstream stage. Additionally, depth information was utilized to augment the 3D object detection~\cite{zhou2022mvsalnet}.  Wang et al.~\cite{wang2019dominant} grouped multiple views with depth information into dominant sets and append the clustered vectors by recurrent cluster-pooling layer. Similar to our approach, \cite{wang2019dominant} leveraged multiple RGBD views for object recognition, achieving improved accuracy compared to single-view methods. However, unlike our approach, its simplistic cluster-pooling strategy fails to effectively capture the unique characteristics of individual viewpoints. Moreover, the proposed method in~\cite{wang2019dominant} added a significant computational burden, ultimately limiting its overall recognition performance.

\section{Methodology}

An overview of the proposed approach is presented in Fig.~\ref{fgvc_overview}. This section provides a detailed explanation of each building block.

\subsection{Pre-Residual Convolutional Encoders}
First, a 3D object is rendered into multiple views, where each view ${v^{j} (j=1, ..., l)}$ includes both RGB and depth images, as illustrated in Fig.~\ref{fgvc_overview}. The resulting multi-modal \textit{RGBD} data from these viewpoints are subsequently processed through convolutional layers, followed by normalization.

After initial processing, the views are fed into the pre-residual convolutional encoders. In the $m$-th pre-residual convolutional stage, the input $\textbf{X}{_{pre}^{j,m-1}}$ undergoes a sequential series of operations, including $\textit{Conv1}{_{pre}^{m}}$, $\textit{BN}{_{pre,1}^{m}}$, $\textit{ReLu}{_{pre,1}^{m}}$, $\textit{Conv2}{_{pre}^{m}}$, and $\textit{BN}{_{pre,2}^{m}}$. Each convolutional operation employs a $3 \times 3$ kernel with a stride of $1$ and padding of $1$. To further enhance residual mapping, the resulting features are combined with $\textbf{X}{_{pre}^{j,m-1}}$, followed by activation through $\textit{ReLu}{_{pre,2}^{m}}$. The processed output is then passed on for subsequent feature extraction in the next stage.

\subsection{Local Transformers}

The local transformer network captures global properties by encoding the output of the pre-residual network. Initially, the output is divided into $p$ patches, which are then projected into sequences of feature vectors, represented as $\mathbf{\textbf{\textit{F}}} = [x_{p_1}^{j}, x_{p_2}^{j}, x_{p_3}^{j}, \ldots, x_{p_w}^{j}]$. Here, $x_{p_w}^{j} \in \mathbb{R}^{192}$ denotes the feature projection of the $p_w$-th patch in the $j$-th view. To incorporate spatial information, a position sequence is added to the feature vectors based on the patch order, along with a \textit{cls\_token} that aggregates global patch information. This results in the input to the local transformer network, formulated as $\mathbf{\textbf{\textit{T}}} = [x_{cls}, x_{p_1}, x_{p_2}, x_{p_3}, \ldots, x_{p_w}] + \mathbf{\textbf{\textit{E}}_{\textbf{\textit{pos}}}}$, where $x_{cls} \in \mathbb{R}^{192}$ is a randomly initialized vector, and $\mathbf{\textbf{\textit{E}}_{\textbf{\textit{pos}}}} = [E_{0}, E_{p_1}, E_{p_2}, E_{p_3}, \ldots, E_{p_w}] \in \mathbb{R}^{(p_w + 1) \times 192}$ denotes the positional embeddings.

The projected embeddings from the pre-residual network, denoted as $\textbf{\textit{T}}_{local}^{j,k-1}$, serve as the input to the local transformer encoders. Each encoder block comprises Layer Normalization (LN), Multi-head Self-Attention (MSA), and a Multi-layer Perceptron (MLP). First, the input $\textbf{\textit{T}}_{local}^{j,k-1}$ undergoes normalization via $\textit{LN}_{local,1}^{k}$ and is processed by $\textit{MSA}_{local}^{k}$, producing an intermediate representation. Subsequently, the intermediate representation, combined with $\textbf{\textit{T}}_{local}^{j,k-1}$,  is passed through $\textit{LN}_{local,2}^{k}$ and $\textit{MLP}_{local}^{k}$, followed by the addition of residual connections. The procedure of the $k$-th local transformer block is illustrated in Fig.~\ref{fgvc_overview}.

\subsection{Middle-Residual Convolutional Encoders}

The middle-residual convolutional network enhances local feature representations derived from transformer-encoded embeddings. As illustrated in Fig.~\ref{fgvc_overview}, the transformer embeddings, excluding the \textit{class\_tokens}, are reshaped into feature maps 
and processed through a series of layers: $\textit{Conv3}{_{mid}^{s}}$, $\textit{BN}{_{mid,1}^{s}}$, $\textit{ReLu}{_{mid,1}^{s}}$, $\textit{Conv4}{_{mid}^{s}}$, $\textit{BN}{_{mid,2}^{s}}$, and $\textit{ReLu}{_{mid,2}^{s}}$.

In the $s$-th encoder of the middle-residual network, the input $\textbf{X}_{mid}^{j,s-1}$ is first passed through $\textit{Conv3}{_{mid}^{s}}$, followed by batch normalization $\textit{BN}{_{mid,1}^{s}}$ and activation using $\textit{ReLu}{_{mid,1}^{s}}$, producing an intermediate feature map $\textbf{X}_{relu_1}^{j,s}$. This intermediate output is further processed by $\textit{Conv4}{_{mid}^{s}}$ and $\textit{BN}{_{mid,2}^{s}}$, after which the residual input $\textbf{X}_{mid}^{j,s-1}$ is added. And then the combined result is activated by $\textit{ReLu}{_{mid,2}^{s}}$.

\subsection{Global Transformers}

The global transformer network is designed to collaboratively process embeddings from multiple views. The synthesized patch features, denoted as $\textbf{T}_{global}^{j,n-1}$, are formed by combining the outputs of middle-residual convolutional blocks and the \textit{class\_tokens} from the local transformer network. This concatenated representation forms the input matrix for the global transformer.

In the $n$-th global transformer encoder, the input $\textbf{T}_{global}^{j,n-1}$ is normalized by $\textit{LN}_{global,1}^{n}$ and then processed by multi-head self-attention ($\textit{MSA}_{global}^{n}$). The output of $\textit{MSA}_{global}^{n}$ is combined with the input $\textbf{T}_{global}^{j,n-1}$ to produce $\textbf{T}_{global,1}^{j,n}$. This intermediate result is further processed by $\textit{LN}_{global,2}^{n}$ and a multi-layer perceptron ($\textit{MLP}_{global}^{n}$), with the output combined with $\textbf{T}_{global,1}^{j,n}$ to generate $\textbf{T}_{global}^{j,n}$, the input for the next stage.

\subsection{Globally Entropy-based Embeddings Fusion (GEEF)}
After processing the input views using global transformers, the object embeddings are represented as $\mathbf{T}_v =[\mathbf{ t}_{v_1}, \mathbf{t}_{v_2}, \ldots, \mathbf{t}_{v_j}, \ldots, \mathbf{t}_{v_l}]$. These extracted representations are then fused and processed through an MLP for object classification. Previous fusion methods commonly use average pooling and max pooling to merge all representations. However, these methods overlook the unique contributions and feature discrepancies across different views, resulting in inadequate fusion of multi-view representations. Moreover, classification embeddings from transformer blocks primarily depend on a single class token, which inadequately captures comprehensive spatial information. To mitigate these limitations, we propose a global entropy-based embedding fusion method to effectively integrate multi-view representations. We begin by applying average pooling to fuse the embeddings of RGB views, denoted as $\mathbf{E}_{v}^{rgb} = [\mathbf{e}_{v_1}, \mathbf{e}_{v_2}, \ldots, \mathbf{e}_{v_j}, \ldots, \mathbf{e}_{v_l}]$, while excluding the class tokens, to obtain the fused representation $\mathbf{E}_{}^{RGB}$.

Next, the entropy of the RGB representation, denoted as $H^{\text{cls}}_v(c_{v_j}) = - \sum_{j=1}^{l} p_{c_{v_j}} \log p_{c_{v_j}}$, is computed from the class token vectors across all RGB views, aggregated in~$\mathbf{C}_{v}^{rgb} = [\mathbf{c}_{v_1}, \mathbf{c}_{v_2}, \ldots, \mathbf{c}_{v_j}, \ldots, \mathbf{c}_{v_l}]$. To utilize entropy values for feature fusion, they need to be normalized to obtain the weights $\mathbf{w}_{v_j}$ for each view. The normalization formula is as follows:
\begin{equation}
w_{v_j} = \frac{H_{v}^{cls}(c_{v_j})}{\sum_{j=1}^{l} H_{v}^{cls}(c_{v_j})}
\end{equation}
where \(w_{v_j}\) represents the weight of the \(j\)-th view. Based on the entropy weights \(w_{v_j}\), a fusion operation of \textit{class\_tokens} is performed on the features from each view. The final fused \textit{class\_tokens} \(\mathbf{C}_{}^{\text{RGB}}\) is expressed as:
\begin{equation}
\mathbf{C}_{}^{\text{RGB}} = \sum_{j=1}^{l} w_{v_j} \mathbf{c}_{v_j}
\end{equation}
Similarly, we compute the corresponding fused depth features, denoted as \(\mathbf{E}_{}^{\text{Depth}}\) and \(\mathbf{C}_{}^{\text{depth}}\). To enhance the representation capability for final classification, a concatenation is applied to aggregate all the representations.

\section{Experiments}

This section summarizes the implementation details and comprehensively evaluates our approach. We analyzed fused-embedding methods, investigated pre- and middle-residual convolutional encoders to determine the optimal LM-MCVT structure, and conducted ablation studies comparing transformers and convolutional encoders. The method's performance and robustness were assessed under varying numbers of views and view structures from multi-modal data. Finally, we compared our results with state-of-the-art methods under consistent viewpoint configurations.

\subsection{ \textit{Implementation Details}}

We used two datasets: the synthetic ModelNet dataset~\cite{wu20153d} and the real-world OmniObject3D dataset~\cite{wu2023omniobject3d}. For the offline experiments on ModelNet, which includes ModelNet10 and ModelNet40, we evaluated our model using the training and testing splits outlined in~\cite{Chen2021MVT}. ModelNet40 contains $12311$ 3D CAD models across 40 categories, with $9843$ models for training and $2468$ for testing. ModelNet10 is a 10-category subset. OmniObject3D is a large-scale dataset of high-quality, real-scanned 3D objects, featuring $6000$ objects across $190$ everyday categories. For OmniObject3D, we adopted the same view settings as ModelNet. To evaluate our approach, we employed a 5-fold cross-validation strategy, which provides a robust estimate of the model's performance and minimizes the risk of overfitting to specific subsets.
We conducted our experiments using an NVIDIA V100 graphics card. Our approach was trained with the Adam optimizer, starting with a learning rate of $0.0002$ and betas set to [$0.9, 0.98$]. To optimize training, we dynamically adjusted the learning rate using a cosine annealing schedule.

\subsection{Performance Analysis of Fusion Strategies}

Our initial experiments explored multi-view fusion techniques. In vision transformer architectures, class token embeddings are commonly used for classification tasks~\cite{dosovitskiy2020image}. However, incorporating embeddings beyond the \textit{class\_tokens} contributes to enhancing the final object representation~\cite{raghu2021vision}. Building on these findings, we proposed the globally entropy-based embedding fusion method (GEEF) and conducted experiments to compare its performance with other fusion techniques.

\begin{table}[!t]
\centering
\vspace{2mm}
\caption{Accuracy (\%) of Different Fused Embedding Methods}
\vspace{-2mm}
\begin{tabular}{ccccc}
\toprule
\multicolumn{1}{c}{\multirow{2}{*}{\textbf{Inputs}}}& \multicolumn{4}{c}{\textbf{Fusion Embedding Methods}}
\\ 
\cmidrule{2-5}
& ACF & ECF & AEF & GEEF \\ \hline
RGB & 94.1 & 93.8 & 93.2  & 94.4 \\
\hline
RGBD& 
94.3& 94.4& 95.4& 96.6\\
\hline
\end{tabular}
\label{tab:fuse_accuracy}

\vspace{-3mm}
\end{table}

\begin{itemize}
\item\textit{Average$-$pooling$-$based Class\_tokens Fusion} (\textbf{ACF}):
This fusion methodology primarily centers on the averaging of class tokens extracted from multiple views~\cite{Chen2021MVT}.

\item\textit{Entropy$-$based Class\_tokens Fusion} (\textbf{ECF}):
ECF fuses multi-view representations using entropy distinctions among class tokens from various views.

\item\textit{Average$-$pooling$-$based Embeddings Fusion} (\textbf{AEF}): This fusion method utilizes average pooling to integrate multiple patch-based image embeddings without including class tokens.
\end{itemize}

To evaluate the effectiveness of GEEF, we conducted 3D object classification experiments on the ModelNet10 dataset. Each object was observed from three different viewpoints with equal interval angles. For a fair comparison, we used the LM-MCVT network, excluding the pre- and middle-residual convolutional encoders.

\begin{figure}[!t]
    \vspace{3mm}
  \centering

    \includegraphics[width=0.65\linewidth]{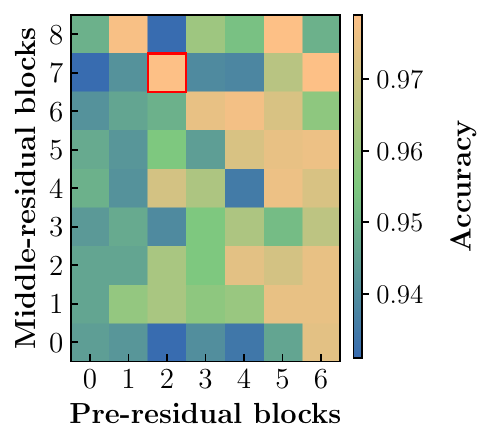}  
    \vspace{-3mm}
  \caption{Ablation study on the effects of pre-residual and middle-residual block configurations on model accuracy.}
    \label{fig:ablation for conv}
    \vspace{-3mm}
\end{figure}

The experimental results in Table~\ref{tab:fuse_accuracy} highlight the superiority of our proposed GEEF method compared to other fusion embedding techniques. GEEF consistently achieves higher accuracy for both RGB and RGBD inputs, outperforming the next-best method (AEF) by up to 1.2\% in the RGBD setting. This improvement can be attributed to GEEF's ability to effectively capture and integrate information from multi-modal multi-view data, leading to a more robust and discriminative feature representation. These findings underscore GEEF's effectiveness in leveraging multi-modal data for enhanced performance.

\subsection{Ablation Study for Convolutional Encoders and Global Transformers}
In this round of experiments, we utilized the ModelNet10 dataset to evaluate the model's performance. We aimed to construct the LM-MCVT network by integrating residual convolutional and transformer encoders. Convolutional encoders excel at extracting local features, while Vision Transformers (ViT) are favored for capturing global information~\cite{raghu2021vision}. However, as the number of ViT layers increases, their capacity to capture local features diminishes~\cite{raghu2021vision}. To address this, we incorporated pre- and middle-residual convolutional layers into the LM-MCVT network to enhance the acquisition of local and global features. Inspired by~\cite{Chen2021MVT}, we employed a local-global structure to facilitate interaction across views. We set local transformers to 8 and global transformers to 4.
We then extensively evaluated the impact of convolutional residual encoders. We analyzed the performance by separately removing pre- and middle-residual blocks and studied the effects of increasing both. The experimental results, illustrated in Fig.~\ref{fig:ablation for conv}, demonstrate that configurations with more convolutional-residual blocks generally achieve better recognition performance. This improvement can be attributed to the enhanced capacity of residual blocks to learn hierarchical and complex feature representations. By mitigating the vanishing gradient problem, residual connections enable deeper networks to be trained effectively without performance degradation, facilitating better extraction and integration of multi-scale features essential for recognizing complex patterns.
LM-MCVT achieves an optimal balance between accuracy and robustness with nine pre- and middle-residual blocks. To ensure a lightweight yet accurate model, we selected a configuration with two pre-residual blocks and seven middle-residual blocks, as show in the red square of Fig.~\ref{fig:ablation for conv}. This finalized network configuration will be used in subsequent experiments.

To further investigate the impact of transformers and convolutional encoders on our model, we conducted an ablation study on each block using multi-modal (RGBD) data. The results, presented in Table~\ref{tab:ablation_study}, emphasize the contributions of each component to the overall performance of LM-MCVT. The full configuration achieves the highest accuracy of 98.0\%, highlighting the synergistic benefits of combining convolutional and transformer-based components. This integration leverages the strengths of both approaches, with convolutional encoders excelling at local feature extraction and transformers capturing long-range dependencies, ultimately leading to optimal performance.

\begin{table}[!t]
    \centering
    \vspace{3mm}

    \caption{Ablation Study on the Impact of Convolutional Encoders and Transformers on Model Accuracy (\%) \\ with Multi-Modal (RGB-D) Data}
    \vspace{-2mm}
    \label{tab:ablation_study}

    \begin{tabular}{cccccc}
        \toprule
        \textbf{Input} & \textbf{\begin{tabular}[c]{@{}c@{}}Without \\ Pre\_cnn\end{tabular}} & 
        \textbf{\begin{tabular}[c]{@{}c@{}}Without \\ LT\end{tabular}} & 
        \textbf{\begin{tabular}[c]{@{}c@{}}Without \\ Mid\_cnn\end{tabular}} & 
        \textbf{\begin{tabular}[c]{@{}c@{}}Without \\ GT\end{tabular}} & 
        \textbf{With All} \\ 
        \midrule
        RGBD  & 94.4 & 96.7 & 95.9 & 96.1 & 98.0 \\  
        \bottomrule
    \end{tabular}

    \vspace{5pt} % 
    \parbox{\textwidth}{
        \footnotesize *LT refers to Local Transformers, and GT refers to Global Transformers.}
        \vspace{-7mm}
\end{table}

\begin{table}[!b]
    \vspace{2mm} % 
    \centering
    \caption{Performance of LM-MCVT on the ModelNet10 Dataset with Different Numbers of Views}
    \label{tab:multimodal_multiview_exps}

    \begin{tabular}{|c|c|cccccc|}
        \hline
        \multicolumn{1}{|c|}{\multirow{2}{*}{\textbf{Inputs}}} & \multicolumn{1}{c|}{\multirow{2}{*}{\textbf{Metric}}} & \multicolumn{6}{c|}{\textbf{Different Number of Views}} \\
        \cline{3-8}
        & & \textbf{1} & \textbf{2} & \textbf{3} & \textbf{4} & \textbf{6} & \textbf{12} \\  
        \hline

        \multirow{2}{*}{\textbf{RGB}} & ACC (\%) & 93.2 & 95.3 & 97.8 & 98.2 & 97.7 & 98.1 \\ 
        \cline{2-8} 
        & Time (ms) & 1.3 & 1.7 & 2.4 & 3.3 & 5.1 & 12.0 \\ 
        \hline

        \multirow{2}{*}{\textbf{RGBD}} & ACC (\%) & 95.5 & 97.2 & 98.0 & 98.5 & 98.1 & 98.9 \\ 
        \cline{2-8} 
        & Time (ms) & 1.7 & 3.2 & 5.1 & 7.0 & 11.6 & 29.7 \\ 
        \hline
    \end{tabular}

    \vspace{5pt} % 
    \parbox{\textwidth}{
        \footnotesize *ACC refers to accuracy, and Time refers to the testing time per instance.
    }

\end{table}

\subsection{Multi-Modal and Multi-View Analysis with Diverse Viewpoint Configurations}

In this section, we present another round of experiments conducted on the ModelNet10 dataset to evaluate the performance of our approach. Multi-modal and multi-view experiments were performed using the viewpoint settings outlined in~\cite{Chen2021MVT}. The results, summarized in Table~\ref{tab:multimodal_multiview_exps}, show that recognition accuracy improves with an increasing number of views but comes at the cost of higher computation time per instance. A four-view configuration was found to offer the optimal balance between accuracy and computational efficiency. While dense views provide more detailed 3D object descriptions, they significantly increase recognition time, making them impractical for real-robot scenarios where computational constraints often limit the number of views robots can capture. This highlights the importance of efficient view selection for balancing performance and practicality.

\begin{figure}[!t]
\centering
\vspace{3.5mm}
\includegraphics[width=0.98\linewidth]{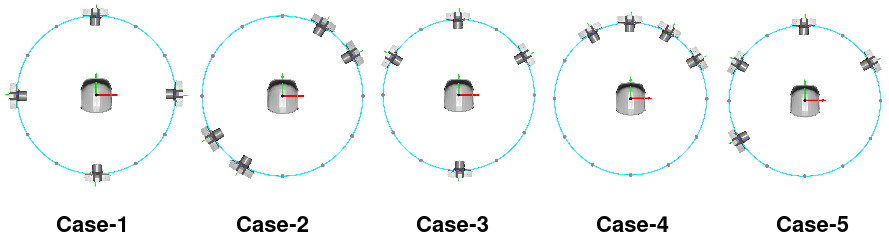}
\caption{Visualization of five cases of four-view structures with circular view-points for the object~(\textit{chair}) in ModelNet10.}
\label{fig:circular_viewpoints}
\end{figure}

In the second phase of our experiments, we evaluated the performance of our method across different structural arrangements of four viewpoints. Towards this goal, five cases of four-view configurations were randomly selected on a circular plane, as illustrated in Fig.~\ref{fig:circular_viewpoints}. The results, summarized in Table~\ref{tab:circular_performance}, demonstrate that our method achieved consistent and reliable performance under diverse viewpoint configurations.
This robustness can be attributed to the method's ability to effectively integrate multi-view information and extract complementary features from different perspectives, ensuring accurate recognition even under varying structural setups.

\begin{figure}[!b]
\centering
\includegraphics[width=0.98\linewidth]{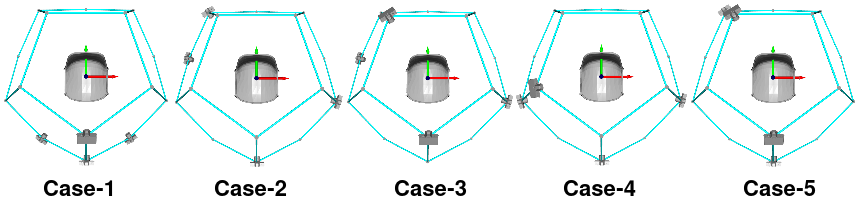}
% \end{tabular}
\caption{Visualization of five cases of four-view structures with hemi-dodecahedron for the object (chair) in ModelNet10.} 
\label{fig:hemispherical}
\end{figure}

Furthermore, we evaluated the performance of our method using four randomly selected viewpoints from hemi-dodecahedron structures. The configurations of these hemi-dodecahedron structures are illustrated in Fig.~\ref{fig:hemispherical}, and the corresponding results are presented in Table~\ref{tab:hemispherical}. These results showed the robustness and effectiveness of our method in handling diverse spatial arrangements. The consistent accuracy across different cases highlights the method's ability to efficiently integrate information from different viewpoints, ensuring reliable performance in 3D recognition tasks.

\subsection{Comparison with State-of-the-art Methods}

\subsubsection{Performance on ModelNet Dataset}

\begin{table}[!t]
\centering
\vspace{-3mm}
\caption{Accuracy (\%) of different circular four-view structures on ModelNet10 dataset}
\resizebox{0.9\linewidth}{!}{
\begin{tabular}{|c|c|c|c|c|c|}
\hline
\multicolumn{1}{|c|}{\multirow{2}{*}{\textbf{Inputs}}}& \multicolumn{5}{c|}{\textbf{Circular four-view structures}}
\\ \cline{2-6}
& \textbf{case-1} & \textbf{case-2 }&\textbf{ case-3 }& \textbf{case-4 }& \textbf{case-5 }\\ \hline
RGB & 98.2 & 95.2 & 97.1 & 96.9 & 96.2 \\

\hline
RGBD& 
98.5& 96.7& 98.2& 96.9& 96.7\\
\hline
\end{tabular}}
\label{tab:circular_performance}
\vspace{-3mm}
\end{table}

% Semi-icosahedral structure

\begin{table}[!b]
\centering
\caption{Accuracy (\%) of different hemi-dodecahedron four-view structures on ModelNet10 dataset}
\resizebox{0.9\linewidth}{!}{
\begin{tabular}{|c|c|c|c|c|c|}
\hline
\multicolumn{1}{|c|}{\multirow{2}{*}{\textbf{Inputs}}}& \multicolumn{5}{c|}{\textbf{ Hemi-dodecahedron four-view structures}}
\\ 
\cline{2-6}
& \textbf{case-1} & \textbf{case-2 }&\textbf{ case-3 }& \textbf{case-4 }& \textbf{case-5 }\\ \hline

RGBD& 
98.2& 99.1& 99.2& 99.0& 98.8\\
\hline
\end{tabular}}
\label{tab:hemispherical}
\end{table}

\begin{table}[!t]
    \vspace{3.5mm}
    \caption{Performance Comparison on \textbf{ModelNet} Dataset}
    \begin{center}
        \resizebox{8.5cm}{4cm}{
        \small
        \resizebox{\linewidth}{!}{
        \begin{tabular}{|c|c|c|c|c|}
        \hline

        \multirow{2}{*}{\textbf{Methods}} & \multirow{2}{*}{\textbf{Params}} & \multirow{2}{*}{\textbf{Views}} & \multicolumn{2}{c|}{\textbf{Accuracy(\%)}} \\
        \cline{4-5}
        &&& \textbf{ModelNet10} & \textbf{ModelNet40} \\

        \hline
        \hline
        \multicolumn{5}{|c|}{Voxel-based Approaches} \\ 
        \hline
        3DShapeNet~\cite{wu20153d}  & - & - & 83.5 & 77.3\\
        VoxNet~\cite{maturana2015voxnet}  & - & - & 92.0& 83.0\\
        VSO~\cite{he2024similarity}  & - & - & 84.0& 72.6\\
        \hline
        \hline
        \multicolumn{5}{|c|}{Point-based Approaches} \\
        \hline
       PCNN~\cite{atzmon2018point}  & 8.1M & - & 94.9& 92.3\\
        PoinTramba~\cite{wang2024pointramba}  & 19.5M & - & -& 92.9\\
       VPC~\cite{gezawa2022voxelized}  & 8.0M & - & 93.4& 88.2\\
       PCM~\cite{zhang2024point}  & 34.2M & - & -& 93.4\\
        \hline
        \hline
        \multicolumn{5}{|c|}{View-based Approaches} \\ 
        \hline
        MVCNN~\cite{su2015multi}  & 103M & 80 & -& 90.1\\
        LSV~\cite{hou2024learning}  & 103M & 12 & -& 94.5\\
        MVTN~\cite{hamdi2024mvtn}  & 73.4M& 12 & -& 92.9\\
        MVT~\cite{Chen2021MVT}  & 22.2M & 12 & 95.3& 94.4\\
        TMTL~\cite{zhang2023tensor}  & 24.0M& 12 & 95.15& 93.68\\ 
        VGP~\cite{han2023viewpoint}  & 11.83M & 12 & 96.47& 95.31\\
        SMV~\cite{Alzahrani_2024_CVPR} & 60.2M & 12 & -& 88.13\\
        \hline
        \hline
        {\textbf{LM-MCVT     (\resizebox{0.6cm}{!}{\textit{ RGB}}}}) &   &  &\textbf{98.2} & \textbf{95.5}\\
        
       {\textbf{LM-MCVT (\resizebox{0.7cm}{!}{\textit{ Depth}}}})  &\textbf{10.5M} &\textbf{4}  &94.1 & 91.9\\  
       
        {\textbf{LM-MCVT(\resizebox{0.7cm}{!}{\textit{ RGBD}}}})  &  &  &\textbf{98.5} & \textbf{95.6}\\
        \hline
        
        \hline
        \end{tabular}}
    }
    \label{summary_tabel}
    \end{center}
    \vspace{-5mm}
\end{table}

As summarized in Table~\ref{summary_tabel}, voxel-based methods, such as 3DShapeNet and VoxNet, show limited performance, achieving a maximum accuracy of 92.0\% on ModelNet10 and 83.0\% on ModelNet40. These approaches struggle with computational efficiency and resolution loss due to the voxelization process, which limits their ability to capture fine-grained details of 3D objects. Point-based methods like PoinTramba and PCNN outperform voxel-based methods by directly operating on point clouds, achieving up to 94.9\% accuracy on ModelNet10. However, these methods still fall short compared to view-based approaches due to challenges in efficiently capturing global and contextual features from sparse point clouds. View-based methods consistently outperform voxel-based and point-based techniques, leveraging multiple 2D projections to better preserve object details and contextual relationships. Among these, MVCNN achieves good accuracy but at the cost of significantly increased computational requirements due to the use of dense multi-view setups.
Our method, LM-MCVT, achieves the best performance while maintaining computational efficiency. Specifically, on ModelNet10, LM-MCVT achieves 98.5\% accuracy using only four views, outperforming heavier methods such as MVCNN, which requires 80 views, and lightweight approaches such as MVT and TMTL, which use 12 views.
On ModelNet40, our method reaches an accuracy of 95.6\%, demonstrating its robustness across datasets. Our method sets a new benchmark in terms of accuracy and efficiency. 
In particular, by effectively combining RGB and depth information, LM-MCVT leverages complementary features from both modalities, providing a richer representation of 3D objects. Furthermore, despite using only 10.5M parameters and four views, our method achieves state-of-the-art performance, offering a practical solution for real-world applications where computational resources and viewpoints are often limited.
We attributed these results to the point that LM-MCVT extracted both local and global features through its convolutional and transformer-based architecture, allowing it to outperform methods with significantly more parameters or views.

\begin{figure*}[!t]
    \vspace{2.5mm}
    \centering
    \includegraphics[width=\linewidth]{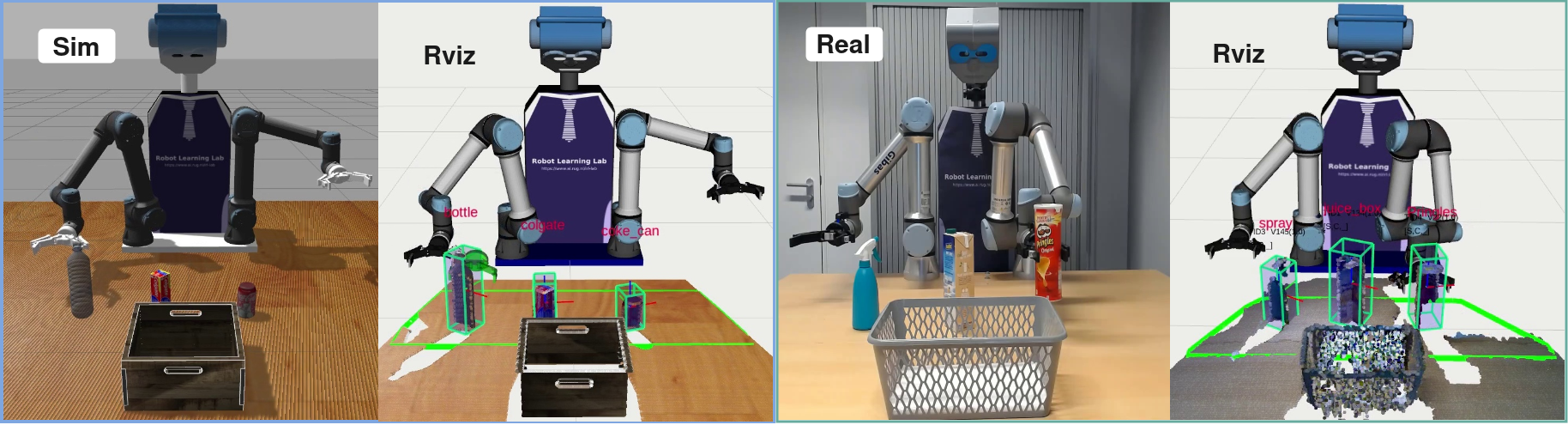}
    \vspace{-5mm}\caption{The snapshots demonstrating the recognition performance of our dual-arm robot employing the LM-MCVT model in~\textit{pick\_and\_place} scenario: (\textit{Sim/Real}) We randomly put three objects in the operational area of the robot. (\textit{Rivz}) The robot is required to successfully recognize these objects using our fine-tuned 4-views LM-MCVT model, before manipulating them. Finally, the robot picked up objects and placed them into the basket.} 

    \label{fig:real-robot}
    \vspace{-2mm}
\end{figure*}

\begin{table}[!t]
\centering
\vspace{4mm}
\caption{Performance Comparison on \textbf{OmniObject3D} Dataset}
\begin{tabular}{|c|c|c|c|c|}
\hline
\textbf{Methods} & \textbf{Input} & \textbf{Params} & \textbf{Accuracy (\%)} & \textbf{Time (ms)} \\ \hline
\multirow{3}{*}{\shortstack{MVT~\cite{Chen2021MVT} \\ (2021)}} 
 & RGB      &   & 60.74  & 6.5  \\ 
 & Depth    & 22.2M  & 62.13  & 6.5  \\ 
 & RGBD     &   & 70.72  & 13.3  \\ \hline
\multirow{3}{*}{\shortstack{SMV~\cite{Alzahrani_2024_CVPR} \\ (2024)}} 
 & RGB      &   & 76.16  & 23.6  \\ 
 & Depth    & 60.2M  & 69.50  & 23.6  \\ 
 & RGBD        &   & 81.02  & 64.3  \\ \hline
\multirow{3}{*}{\shortstack{\textbf{LM-MCVT} \\ (Ours)}} 
 & RGB      &   & \textbf{83.33}  &\textbf{3.3}  \\ 
 & Depth    & 10.5M  & 78.80  &\textbf{3.3}  \\ 
 & RGBD        &   &\textbf{ 85.10}  &\textbf{7.0 } \\ \hline
\end{tabular}
\label{tab:omniobject3d}
\vspace{-5mm}
\end{table}
\subsubsection{Performance on OmniObject3D Dataset}~In this round of experiments, we conducted 5-fold cross-validation using four views. Since there are no established baselines for the OmniObject3D dataset, we trained all multi-view methods from scratch. The experimental results, presented in Table~\ref{tab:omniobject3d}, demonstrate the performance of various methods on this challenging dataset. The OmniObject3D dataset is significantly more complex than ModelNet40. This increased complexity resulted in a performance drop across all methods compared to the ModelNet40 experiments. Despite this, our method, LM-MCVT, achieved a remarkable accuracy of 85.1\%, outperforming the competing approaches. Specifically, LM-MCVT demonstrated superior performance across all input modalities (RGB, Depth, and RGBD), with its RGBD accuracy significantly surpassing that of SMV (81.0\%) and MVT (70.7\%).
In addition to its accuracy, LM-MCVT proved to be highly computationally efficient. With an inference time of only 7.0 ms for RGBD inputs, it outperformed MVT (13.3 ms) and SMV (64.3 ms) while utilizing much fewer parameters (10.5M compared to 22.2M for MVT and 60.2M for SMV). This efficiency highlights the lightweight nature of our method, making it suitable for real-time applications.
Moreover, LM-MCVT effectively leverages multi-modal data, maintaining high accuracy across different input types. This capability is particularly advantageous for the OmniObject3D dataset, which has diverse object categories requiring robust feature integration.

\subsection{Robotic Demonstrations}

To evaluate our method in a robotic setting, we initially fine-tuned the pre-trained LM-MCVT model on the Synthetic Household Object dataset~\cite{kasaei2024lifelong}. For each object, we captured four views. This fine-tuning was necessary because the objects are partially visible to the robot due to (self) occlusion. The LM-MCVT model was fine-tuned using a 5-fold cross-validation approach, achieving a recognition accuracy of $99.3\%$ on Synthetic Household Objects. In the experiment, we randomly positioned three objects in front of the robot, as depicted in Fig.~\ref{fig:real-robot}. The first step involved segmenting the object from the surrounding environment~\cite{kasaei2018towards, kasaei2020orthographicnet}. The robot then captured four RGBD views of the segmented objects, as illustrated in Fig.~\ref{fgvc_overview}. These views were fed into the fine-tuned LM-MCVT model to identify the object. Subsequently, the robot detected a suitable grasp configuration for each target object, grasping and relocating them into a basket \cite{kasaei2023mvgrasp}. The process is detailed in Fig.~\ref{fig:real-robot}, which shows the performance of the robot in sim-real settings and the visualization of recognition in RViz. We conducted ten rounds of experiments. 
Across all trials, the robot consistently identified, picked up, and placed the objects correctly, demonstrating the robustness and reliability of the LM-MCVT model in handling real-world variations. The success of these experiments underscores the practical applicability of our model in real-robot settings.

\section{Conclusion}

In this study, we introduced the Lightweight Multi-Modal Multi-View Convolutional-Vision Transformer (LM-MCVT) network, an innovative framework designed to enhance 3D object recognition in various real-world applications. Our approach integrates convolutional encoders and transformers to effectively extract and fuse multi-modal, multi-view data. A key contribution of our work is the introduction of the Globally Entropy-based Embeddings Fusion (GEEF) method, which optimally combines information from different views to improve recognition accuracy. Through a series of comprehensive experiments, we demonstrated the superiority of the LM-MCVT network in comparison to existing state-of-the-art methods. Notably, our model achieved state-of-the-art performance on both synthetic and real-world datasets, including the ModelNet and OmniObject3D datasets. Specifically, the LM-MCVT network exhibited superior accuracy while maintaining a lightweight architecture, making it suitable for deployment in environments with computational constraints. Moreover, the integration of the LM-MCVT model into a dual-arm robotic framework highlighted its practical applicability. The robot was able to accurately recognize and manipulate objects, demonstrating the model's robustness and effectiveness. In the continuation of this work, inspired by the advancements in large vision-language models (VLMs), such as those demonstrated in \cite{tziafas2024towards}, we plan to investigate a hybrid approach that combines geometric and visual information with the contextual reasoning capabilities of VLMs.

\section*{ACKNOWLEDGMENT}
We thank the Center for Information Technology of the University of Groningen for their support and for providing access to the high performance computing cluster - Hábrók.

\bibliographystyle{IEEEtran}
\small\bibliography{reference}

% %%%%%%%%%%%%%%%%%%%%%%%%%%%%%%%%%%%%%%%%%%%%%%%%%%%%%%%%%%%%%%%%%%%%%%%%%%%%%%%%

% %%%%%%%%%%%%%%%%%%%%%%%%%%%%%%%%%%%%%%%%%%%%%%%%%%%%%%%%%%%%%%%%%%%%%%%%%%%%%%%%

% %%%%%%%%%%%%%%%%%%%%%%%%%%%%%%%%%%%%%%%%%%%%%%%%%%%%%%%%%%%%%%%%%%%%%%%%%%%%%%%%
% \section*{APPENDIX}

% Appendixes should appear before the acknowledgment.

%%%%%%%%%%%%%%%%%%%%%%%%%%%%%%%%%%%%%%%%%%%%%%%%%%%%%%%%%%%%%%%%%%%%%%%%%%%%%%%%

\end{document}